# Predicting customer lifetime value – e-commerce use case

Ziv Pollak
Machine Learning Eng.
Ziv.Pollak@clearly.ca

Predicting customer future purchases and lifetime value is a key metrics for managing marketing campaigns and optimizing marketing spend. This task is specifically challenging when the relationships between the customer and the firm are of a noncontractual nature and therefore the future purchases need to be predicted based mostly on historical purchases.

This work compares two approaches to predict customer future purchases, first using a "buy-till-you-die" statistical model to predict customer behavior and later using a neural network on the same dataset and comparing the results. This comparison will lead to both quantitative and qualitative analysis of those two methods as well as recommendation on how to proceed in different cases and opportunities for future research.

1. **Introduction**

Customer lifetime value (LTV) is a key metrics for managing marketing campaigns, as described in [1], [2], [3], [4] and [5]. One of the main challenges is accurately predicting future purchase behavior when the relationships between the customer and the firm are of a noncontractual nature [6], [7]. There are 2 main challenges in this scenario:

- First, in such a setting, a customer's current status is not directly observable by the organization but must be inferred indirectly from past activity.
- Second, the amount of customer-level data tends to vary significantly. In general, only very few transactions are observed for most customers, and hence to extract the most information from the available data, marketing analysts need to adaptively pool information across customers.

This work will start by reviewing both a statistics-based solution and a machine learning solutions to the problem of predicting future purchases. Later, we will move into comparing the actual implementation of those models on a real-life e-commerce dataset with about 2 million customers. The results of those implementations will be analyzed and compared. To close, we will review possible direction for future progress.

## 2. Related work

Statistical models of customer purchasing behavior have been studied for many years. Early models were often restricted to simple parametric statistical models given the limited amount of compute and data that were available. With the progress made on e-commerce platforms this problem began to attract the attention of machine learning researchers as well.

### 2.1 Distribution fitting approach

A series of probabilistic models link simple past purchase summary statistics with a theoretically well-grounded behavioral [8]. These so-called "buy-till-you-die" (BTYD) models represent the well-established standard approach used by data scientists and analysts to predict LTV components (such as the expected number of future purchases) and have been applied in a wide variety of industries [9].

The most widely recognized benchmark BTYD model, is the Pareto/NBD [10] model which assumes a Poisson purchase process and an exponentially distributed lifetime. The two-corresponding customer-level parameters can vary across customers, following independent gamma distributions. Because the gamma-exponential mixture results in a Pareto distribution, whereas the gamma-Poisson mixture leads to a negative binomial distribution, the model is referred to as Pareto/NBD.

The drawback of these "buy-till-you-die" models is that those models interpret the purchase hiatus at the end of the observation period equally, however, their observed inter-transaction timing patterns tell a different story. A model that adequately accounts for these differences in timing patterns would thus assign a significantly lower probability of future activities to the regular but "overdue" customer relative to the customer purchasing more irregularly.

[11] developed a Pareto/NBD variant that replaces the NBD repeat-buying component with a mixture of gamma distributions (named Pareto/GGG) to allow for a varying degree of regularity across customers. This new model, who accounts for regularity, outperforms both the Pareto/NBD and the heuristic benchmark suggested by [9] in terms of out-of-sample, individual-level forecasting accuracy for future customer-level transactions. This is already the case for relatively mild inter-transaction timing regularities, whereas the models perform on par for data sets without regularities. The greatest improvements accrue in the important, high-frequency, low-recency customer segment. In such a situation, a standard NBD-type model results in overly optimistic predictions and erroneous recommendations with respect to customer prioritization.

## 2.2 Machine learning approach

More recent works on LTV prediction are using machine learning models to predict customer purchases and lifetime value. [12] employed random forests to estimate the LTV of the costumers of an online fashion retailer, while [13], is the first work to explicitly include customer engagement features in LTV prediction model. They also address the challenges of learning the complex, non-linear LTV distribution by solving several simpler sub-problems. Firstly, a binary classifier is run to identify customers with predicted CLTV greater than zero. Secondly, the customers predicted to shop again are split into five groups and independent regressors are trained for each group.

## 3. Use case details

### 3.1 Dataset

To compare the Pareto/GGG model and the neural network implementation we have used Clearly's customer purchase historical data (www.clearly.ca), which includes about 2 million customers from the last 5 years.

The following features were extracted for each customer:

| Name | Description |
| --- | --- |
| Lifetime_duration | Days from first buy until last buy |
| num_purchases | Number of historical purchases |
| avg_gaps | Average days between purchases |
| avg_revenue | Average revenue from purchases |
| days_ago_first_buy | Days since first buy |
| days_ago_last_buy | Days since last buy |
| num_coupons | How many coupons used by the customer |

Our goal was to try and predict customer purchases in the next 6 months. For that we have used the data from June 2015 and until June 2020 to train the model, and data from June 2020 to Dec 2020 to test the model (e.g. a 6 month holdout period)

### 3.2 Model implementation

The statistical model implementation was based on the purchase history data, using the detailed purchase date and purchase amount data. We have leveraged the Lifetimes Python library [14] when implementing the model. It was clear, even before deep diving into the results, that we can achieve reasonably good prediction very fast.

On the other hand, we have implemented a fully connected neural network, utilizing the summarized data described above. While building the neural network model, we have tested multiple different configurations and ended up with the one that is described in the table below:

| Layer | Type | Size | Dropout | Activation |
|---|---|---|---|---|
| Layer#1 | Dense | 128 | 0 | Relu |
| Layer#2 | Dense | 256 | 0 | Relu |
| Layer#3 | Dense | 512 | 0 | Relu |
| Layer#4 | Dense | 32 | 0 | Relu |
| Layer#5 | Dense | 1 | 0 | Relu |

4. **Performance evaluation and analysis**

When comparing the results of the statistical model vs the NN model, we reviewed both quantitative and qualitative benefits of it.

    4.1 Quantitative comparison

When looking at quantitative benefits, using the statistical mode has some major advantages:

- The biggest advantage of the statistical model is it relatively simple and fast to compute. There are no long training efforts needed.
- The data used by the model is simple and no feature engineering is needed, it is basically just detailed purchase date and purchase amount data.
- The model can produce the "probably alive" for each customer, which can be used to guide marketing decisions.
- Explain-ability – the model produces expected purchase count as well as purchase amount which is easy to explain.

On the other hand, the advantages of the neural network model are:

- Ability to use additional data fields, which can include more details on the products or the customer.

- Quickly adjust to market changes – When market condition change it is easy to re-train the model based on the latest data.

### 4.2 Qualitative comparison

When looking at the qualitative results comparison we chose to focus on only one aspect – the accuracy of the prediction when compared to the real results. This parameter is simple to compare and give us a clear indication on the usability of the model for marketing needs (as described above).

The accuracy of each model is described in the table below:

|  | Pareto/GGG | Neural Network |
|---|---|---|
| % Accuracy on hold-out period | 88.6% | 94.6% |

It is clear, that the neural network model produced much better prediction when compared to the statistical model. This can be attributed to the different input coming into the model as well as the ability of the neural network to learn complex functions.

### 4.3 Model comparison summary

To sum, the selection of using a statistical model vs a neural network model will depend on the uses of the data coming from the model and the needed accuracy.

If explain-ability, "probability alive" or fast processing time are important the statistical model is a good way to solve this problem. When trying to get the highest accuracy level, even if it requires more efforts with feature engineering and more processing time, then the neural network is the way to go.

## 5. Conclusion & Future work

This work provides details analysis of real use case comparing the calculation of customer lifetime value (LTV) using real data on both a statistical and a neural network model. We have compared the two models and analyzed the advantages and benefits of each.

We have demonstrated that neural network can produce better prediction results, but it come with a price of both data processing effort and training time and compute resources.

The comparison between the statistical model and the neural network highlight 2 potential directions for future work: First, there is room to improve the statistical models and adjust it the specific conditions we see in the data, to allow it to achieve better results. Second, given the long time and effort needed to train the neural network model we need to investigate ways to reduce this effort (possible using transfer learning).